\documentclass[11pt]{article}

\usepackage[preprint]{acl}

\usepackage{times}
\usepackage{latexsym}
\usepackage{fontawesome5}
 \usepackage{amsmath} 
 \usepackage{flafter}
 \usepackage{wrapfig}
\usepackage{amsthm}
\usepackage{enumitem}
\usepackage{amssymb}
\newtheorem{proposition}{Proposition}[section]
\usepackage{booktabs}
\usepackage{graphicx}
\usepackage{capt-of}
\usepackage{booktabs}
\usepackage{xcolor}

\definecolor{gaingreen}{RGB}{0,110,70}
\newcommand{\gain}[1]{%
  \hspace{0.5pt}%
  \raisebox{-0.3ex}{%
    \textcolor{gaingreen}{%
      \fontsize{8}{9}\selectfont\bfseries +#1%
    }%
  }%
}

\usepackage[T1]{fontenc}

\usepackage[utf8]{inputenc}

\usepackage{microtype}

\usepackage{inconsolata}

\usepackage{graphicx}
\usepackage{float}
\usepackage{cuted}
\usepackage{capt-of}

\title{Beyond the Teacher: Understanding and Preserving Distillation Overshoot in Agentic On-Policy Distillation}

\newsavebox{\spodAuthorRow}
\newsavebox{\spodAuthorCandidate}
\newcommand{\spodStartRows}{\sbox{\spodAuthorRow}{}}
\newcommand{\spodAddEntry}[1]{%
  \ifdim\wd\spodAuthorRow=0pt
    \sbox{\spodAuthorRow}{#1}%
  \else
    \sbox{\spodAuthorCandidate}{\usebox{\spodAuthorRow}\hspace{0.8em}#1}%
    \ifdim\wd\spodAuthorCandidate>\linewidth
      \usebox{\spodAuthorRow}\par
      \sbox{\spodAuthorRow}{#1}%
    \else
      \sbox{\spodAuthorRow}{\usebox{\spodAuthorCandidate}}%
    \fi
  \fi
}
\newcommand{\spodEndRows}{\usebox{\spodAuthorRow}\par}
\author{%
\begin{minipage}{\dimexpr\textwidth-2\tabcolsep\relax}
\centering\large\bfseries
\spodStartRows
\spodAddEntry{Jike Zhong\textsuperscript{1*},}
\spodAddEntry{Ming Li\textsuperscript{2*},}
\spodAddEntry{Yuxiang Lai\textsuperscript{3},}
\spodAddEntry{Xi Xiao\textsuperscript{4},}
\spodAddEntry{Sheng Zhang\textsuperscript{5},}
\spodAddEntry{Ritwick Chaudhry\textsuperscript{6},}
\spodAddEntry{Tianle Chen\textsuperscript{7},}
\spodAddEntry{Jingyu Xie\textsuperscript{1},}
\spodAddEntry{Zheda Mai\textsuperscript{8},}
\spodAddEntry{Shao-Yuan Lo\textsuperscript{9},}
\spodAddEntry{Konstantinos Psounis\textsuperscript{1},}
\spodAddEntry{Yuanyuan Lei\textsuperscript{2}}
\spodEndRows
\vspace{6pt}
\small\normalfont
\spodStartRows
\spodAddEntry{\textsuperscript{1}University of Southern California}
\spodAddEntry{\textsuperscript{2}University of Florida}
\spodAddEntry{\textsuperscript{3}Emory University}
\spodAddEntry{\textsuperscript{4}University of Alabama}
\spodAddEntry{\textsuperscript{5}University of Maryland}
\spodAddEntry{\textsuperscript{6}Amazon AGI}
\spodAddEntry{\textsuperscript{7}Boston University}
\spodAddEntry{\textsuperscript{8}The Ohio State University}
\spodAddEntry{\textsuperscript{9}National Taiwan University}
\spodEndRows
\vspace{3pt}
\href{https://github.com/Jike338/SPOD}{\faGithub\enspace\textbf{https://github.com/Jike338/SPOD}}
\end{minipage}%
}

\begin{document}
\raggedbottom

\makeatletter
\g@addto@macro\@maketitle{%
  \par
  \vspace{-12pt}
  \begin{minipage}{\textwidth}
  
    \nolinenumbers
    \centering
    \includegraphics[width=0.96\linewidth]
      {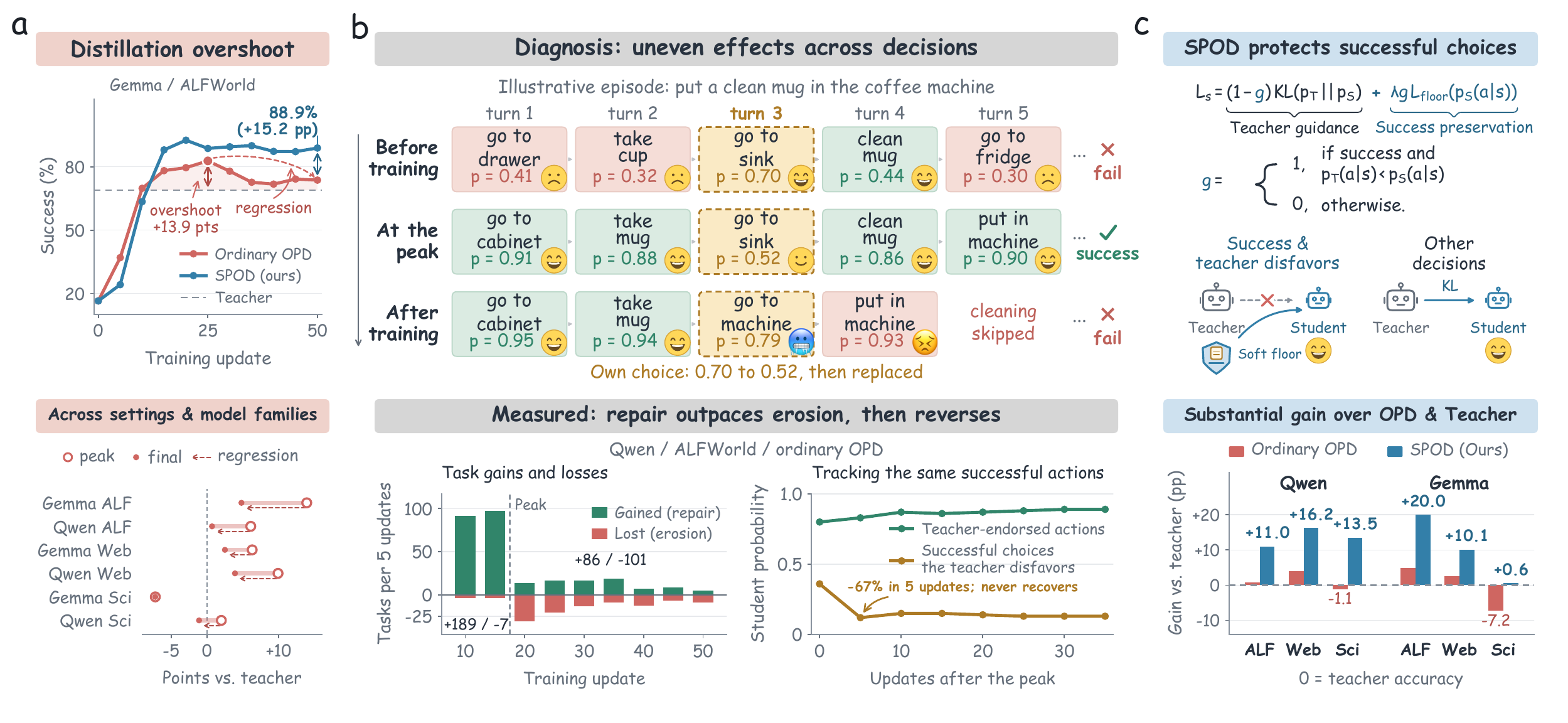}
    \captionof{figure}{%
      \textbf{Understanding and preserving distillation overshoot.}
      (a) Ordinary OPD students can substantially surpass their
      teachers mid-training, but then regress as training continues.
      (b) OPD affects decisions unevenly: repair of the student's
      weak decisions initially outpaces erosion of its own
      successful choices, producing overshoot; continued
      distillation then erodes those choices, driving regression.
      (c) SPOD withholds conflicting supervision on successful
      choices, preserving and extending gains in all settings,
      effectively curbing regression.}
    \label{fig:teaser}
  \end{minipage}
  \vspace{13pt}
}
\makeatother

\maketitle
\begingroup
\renewcommand{\thefootnote}{*}
\footnotetext{Equal contribution.}
\endgroup

\begin{abstract}

On-policy distillation (OPD) trains agents on their own trajectories using supervision from a stronger teacher. We uncover a counterintuitive phenomenon in multi-turn agentic tasks: a smaller, initially weaker student can substantially outperform its teacher during training, only to regress as distillation continues. We term this phenomenon \textbf{\textit{distillation overshoot}}. Through decision- and task-level analyses, we trace overshoot to two competing effects of OPD: repairing weak student decisions and eroding successful student choices that the teacher disfavors. Early in training, repair outpaces erosion, allowing the student to combine the teacher's corrections with its own strengths and surpass the teacher. As training continues, erosion overtakes repair, causing these gains to vanish. Motivated by this finding, we propose \textbf{S}uccess-\textbf{P}reserving \textbf{O}n-Policy \textbf{D}istillation (\textbf{SPOD}), which selectively masks conflicting teacher supervision on successful student choices and introduces a soft probability floor to counteract their erosion, while retaining teacher guidance elsewhere. SPOD requires only binary episode success and model probabilities already available during OPD. Across two model families and three agentic benchmarks, SPOD consistently outperforms ordinary OPD and evaluated variants, improving final success by up to 15.2 percentage points and surpassing the teacher by up to 20.0 points. Moreover, SPOD's final checkpoints outperform even ordinary OPD's best intermediate checkpoints in all six settings, demonstrating that beyond-teacher gains can be not only preserved but further extended.

\end{abstract}

\section{Introduction}
\label{sec:intro}

On-policy distillation (OPD) has become an effective approach for training smaller language models under the guidance of larger teachers \citep{agarwal2024gkd,gu2024minillm,qwen3,lu2025opd}, and is increasingly adopted for multi-turn agentic tasks \citep{wang2026tcod,li2026guidedopd,zhou2026sageopd,zhong2026sod}. By matching the teacher's action distribution at each decision state along the student's own trajectories, OPD reduces the train-test mismatch associated with fixed demonstrations \citep{ross2011dagger}. Since the teacher serves as the target of distillation \citep{hinton2015distilling,gou2021kdsurvey}, recent approaches have sought to surpass its performance through stronger supervision or additional reinforcement learning \citep{wang2026madopd,tan2026atod,ye2026opdsearch,yu2026retireopd}. 

\begin{figure}
    \centering
    \includegraphics[width=1\linewidth]{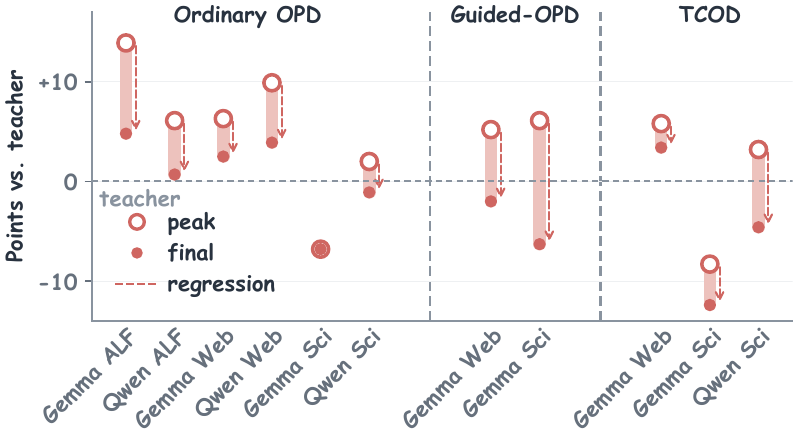}
\caption{\textbf{Overshoot-then-regression appears across multiple OPD methods (9 out of 11 runs).}}
    \label{fig:overshoot widely exists}
    \vspace{-10pt}
\end{figure}

In this work, we uncover a \textbf{\textit{distillation overshoot}} phenomenon in agentic OPD: a smaller, initially much weaker student can substantially outperform its larger teacher during training, only to lose much of this advantage as distillation continues. For instance, on ALFWorld, a Gemma E4B student improves from an initial 16.5\% success, compared with 68.9\% for its 31B teacher, to a peak of 82.8\%, before declining to 73.7\% by the end of training (\autoref{fig:teaser}(a)). Similar overshoot-then-regression patterns emerge across multiple model--benchmark settings and OPD variants (\autoref{fig:overshoot widely exists}). These findings challenge the view of teacher performance as a natural ceiling for the distillation process.

\begin{figure}[!t]
    \centering
    \includegraphics[width=\linewidth]{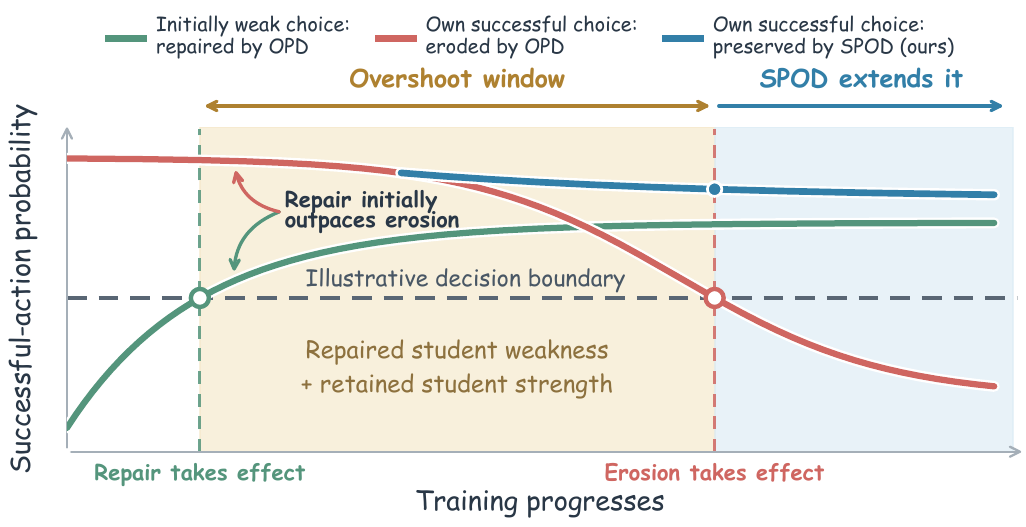}
\caption{Schematic illustration of the temporary overshoot ``window effect'' in agentic OPD.}    
\label{fig:overshoot_window}
\vspace{-10pt}
\end{figure}

To understand this overshoot phenomenon, we examine how OPD affects individual student decisions and their impact on task success. Our analysis reveals two competing effects: \emph{repair} of weak student decisions and \emph{erosion} of successful student choices that the teacher disfavors (\autoref{fig:overshoot_window}). Crucially, these effects unfold at different rates. Early in training, repair outpaces erosion, allowing the student to combine teacher corrections with its own strengths and surpass the teacher. As distillation continues, however, erosion overtakes repair, driving regression. We substantiate this mechanism at both the decision and task levels. Successful actions strongly disfavored by the teacher lose about two-thirds of their probability within five updates and remain suppressed, while teacher-endorsed actions continue to strengthen (\autoref{fig:teaser}(b), bottom right). At the task level, gains dominate losses before the peak, but this balance reverses afterward (\autoref{fig:teaser}(b), bottom left). 

This diagnosis suggests that preserving beyond-teacher gains requires protecting successful student choices while retaining useful teacher guidance. Existing approaches fall short: selecting an intermediate checkpoint halts further learning and cannot preserve gains that peak at different times across tasks, while retiring the teacher \citep{yu2026retireopd} also removes potentially beneficial corrections. Recent agentic OPD methods adjust rollout horizons, teacher interventions, supervision selection, or loss weights \citep{wang2026tcod,li2026guidedopd,zhou2026sageopd,zhong2026sod}, but do not explicitly protect successful student choices from conflicting teacher supervision. 

Motivated by this diagnosis, we propose \textbf{S}uccess-\textbf{P}reserving \textbf{O}n-Policy \textbf{D}istillation (\textbf{SPOD}), a simple modification of OPD that preserves successful student choices while continuing to learn from the teacher elsewhere. Specifically, when an action belongs to a successful student trajectory and the teacher assigns it lower probability than the student, SPOD masks the corresponding distillation loss and applies a soft probability floor. These two components address complementary sources of erosion: masking removes the direct gradient that pulls the student toward the teacher, while the probability floor counteracts indirect suppression caused by parameter updates at other decisions. All remaining actions receive ordinary teacher supervision (\autoref{fig:teaser}(c)). SPOD uses only binary episode success and model probabilities already available in OPD, requiring no additional reward model, replay buffer, or task-family-specific supervision.

Experiments over two model families (Qwen \citep{qwen35} and Gemma \citep{gemma4}) and three multi-turn agentic benchmarks (ALFWorld \citep{shridhar2021alfworld}, WebShop \citep{yao2022webshop}, and ScienceWorld \citep{wang2022scienceworld}) demonstrate that SPOD outperforms ordinary OPD, the teacher, and evaluated OPD variants, improving average final success over ordinary OPD by up to 12.4 points. Gains reach 15.2 points over ordinary OPD and 20.0 points over the teacher. Notably, SPOD's final checkpoints surpass even ordinary OPD's best intermediate checkpoints in every setting, demonstrating that it not only preserves but also extends the overshoot gains. Ablation studies further confirm the complementary roles of masking conflicting supervision and preserving successful action probabilities. To summarize, our contributions are:
\begin{itemize}
    \item We reveal \emph{distillation overshoot} phenomenon in agentic OPD: smaller, initially weaker students can substantially surpass their larger teachers mid-training before regressing.
    \item We trace this phenomenon to the competing effects of \textit{repairing} weak student decisions and \textit{eroding} successful choices that the teacher disfavors, explaining the beyond-teacher performance and its subsequent decline.
    \item We propose SPOD, a distillation objective that selectively protects successful student choices from conflicting teacher supervision while retaining useful guidance, surpassing the best intermediate checkpoints of ordinary OPD.
\end{itemize}

\section{Related Work}
\label{sec:related}

\paragraph{On-policy distillation.}
OPD trains students on their own visited states, reducing
the train--deployment mismatch of fixed demonstrations
\citep{ross2011dagger}. GKD applies this principle to language
models \citep{agarwal2024gkd}, MiniLLM uses reverse KL
\citep{gu2024minillm}, and OPD is widely used in model
post-training \citep{qwen3,lu2025opd}. For multi-turn agents,
TCOD schedules rollout horizons \citep{wang2026tcod},
Guided-OPD introduces teacher turns \citep{li2026guidedopd},
SAGE-OPD selects interventions using teacher judgments
\citep{zhou2026sageopd}, and SOD and ATOD reweight supervision
using student--teacher disagreement
\citep{zhong2026sod,tan2026atod}. Less is More truncates
rollouts to avoid degraded teacher guidance at later token
positions \citep{zhou2026less}. These methods adjust the
collection or application of teacher guidance while SPOD
selectively protects successful student choices that
this guidance would erode.

\paragraph{Students surpassing teachers.}
Students can exceed their teachers through same-size
distillation \citep{furlanello2018born,agarwal2024gkd},
weak-to-strong learning \citep{burns2024weak,yu2026w2sopd},
and multi-teacher supervision \citep{wang2026madopd}.
OPD studies likewise report students outperforming
their teachers after training
\citep{wang2026tcod,tan2026atod,yu2026retireopd,nie2026states}.
However, we show that final performance above the teacher
can conceal a much stronger intermediate policy, even for
smaller, initially weaker students in multi-turn agentic OPD.
We further explain this rise and regression through competing
repair and erosion, motivating SPOD.

\section{Understanding Distillation Overshoot}
\label{sec:analysis}

Having established the overshoot pattern (\autoref{fig:teaser} \& \autoref{fig:overshoot widely exists} ), we investigate how it occurs and why continued distillation erodes this advantage.

\subsection{Analysis Setup}

We study ordinary forward-KL OPD on ALFWorld using
Qwen3.5-2B/27B and Gemma-4-E4B/31B student--teacher pairs.
Given an interaction history $s$, the student $p_\theta$
generates an action $a=(a_1,\ldots,a_n)$ and learns from
a frozen teacher $q$ through the token-averaged OPD loss
\begin{equation}
 \ell_{\mathrm{KD}}(s,a)=\frac{1}{n}\sum_{j=1}^{n}
 D_{\mathrm{KL}}(q_j\Vert p_{\theta,j}).
 \label{eq:fkl}
\end{equation}
Both distributions condition on the same history and
student-generated prefix $(s,a_{<j})$, using each model's
native prompt template. Sampled histories and tokens
are held fixed during differentiation.

Our analysis links decision-level changes to task-level
outcomes. Every five updates, we rescore fixed state--action
pairs to track decision probabilities and evaluate success
on 274 seen/unseen tasks, averaging success over three
evaluation seeds. \autoref{sec:experiments} evaluates SPOD objective
across all benchmarks and model families without re-tuning.

\begin{figure}[t]
    \centering
    \includegraphics[width=0.95\linewidth]{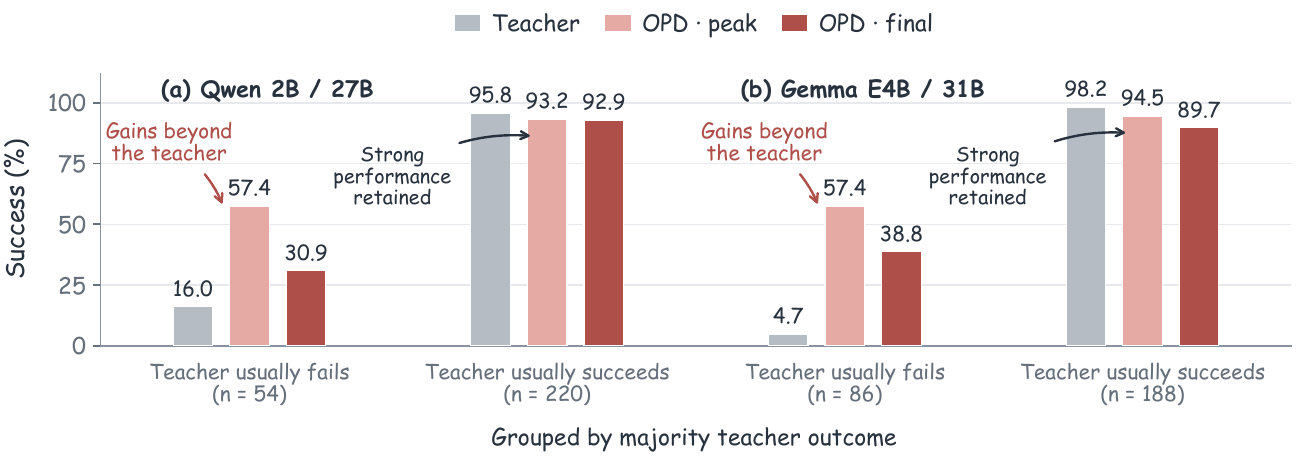}
\caption{\textbf{Where the overshoot comes from.}
At its peak, ordinary OPD substantially outperforms the teacher
on tasks the teacher usually fails while retaining strong
performance on tasks it usually solves. Subsequent regression
is concentrated in the former group. Tasks are grouped by
majority teacher outcome on ALFWorld, and $n$ denotes cohort size.}
    \label{fig:teacher_cohorts}
    \vspace{-10pt}
\end{figure}

\subsection{Overshoot Emerges as Repair Initially Outpaces Erosion}
\label{sec:emergence}

\paragraph{Repair initially outpaces erosion within trajectories.}
To understand why the student surpasses its teacher, we first examine
individual decisions. In a successfully learned Qwen episode (``heat some apple and
put it in sidetable''), the probability of ``go to microwave'' rises from
$9.3\times10^{-5}$ to 0.87 by update 15, completing the failure-to-success flip, even though the already-correct ``heat apple'' falls only from 0.82 to 0.58
(teacher: 0.076); a second example appears in \autoref{fig:teaser}b (top).
The pattern is general: 82.7\% of strongly teacher-disfavored actions from
successful episodes at update 11 halve by update 15, yet task success still rises from
52.1\% to 86.1\%. Thus early in training, repair often outpaces erosion.

\paragraph{Decision-level imbalance yields beyond-teacher task-level performance.}
Consistent with this mechanism, task-level gains initially far exceed
losses. Summing increases and decreases in per-task success between
adjacent checkpoints, Qwen gains 188.7 task-equivalents from update 5 to
its peak at update 15 but loses only 7.3 (\autoref{fig:teaser}b(bottom left)); gains
exceed losses before the peak in all five ordinary-OPD runs across both
model families. Moreover, at the peak both students solve 57.4\% of the
ALFWorld tasks the teacher usually fails (teachers: 16.0\% and 4.7\%),
while still solving 93.2\% and 94.5\% of those it usually solves
(\autoref{fig:teacher_cohorts}). Consequently, the peak student surpasses
the teacher by \textbf{retaining its reliable behavior while adding successes
where it is unreliable}.

\subsection{Regression Emerges as Erosion Overtakes Repair}
\label{sec:erosion}

\paragraph{Continued imitation suppresses successful student choices.}
Despite repair's initial lead, erosion still occurs, and its source is the distillation objective itself. Specifically, at a token position, forward KL has the exact logit derivative
\begin{equation}
 \frac{\partial D_{\mathrm{KL}}(q\Vert p_\theta)}{\partial z_v}
 = p_\theta(v)-q(v).
 \label{eq:logit}
\end{equation}
Whenever the student assigns a token higher probability than the
teacher, this gradient pushes its logit down, regardless of whether
the episode succeeded. Ordinary OPD thus treats a successful departure
from the teacher the same as a mistake, and this pressure accumulates
on the student's useful choices as training continues.

\paragraph{Successful choices erode as teacher-supported actions strengthen.}
We examine this pressure empirically by rescoring fixed state--action
pairs from successful trajectories near the peak. In Qwen,
teacher-supported actions rise in mean probability from 0.80 to 0.89,
whereas strongly teacher-disfavored successful actions fall by 67\%,
from 0.36 to 0.12, within five updates and remain low
(\autoref{fig:teaser}b(bottom right)). Continued imitation thus strengthens
teacher-supported behavior while eroding the student's useful
departures from it.

\begin{figure}
    \centering
    \includegraphics[width=0.95\linewidth]{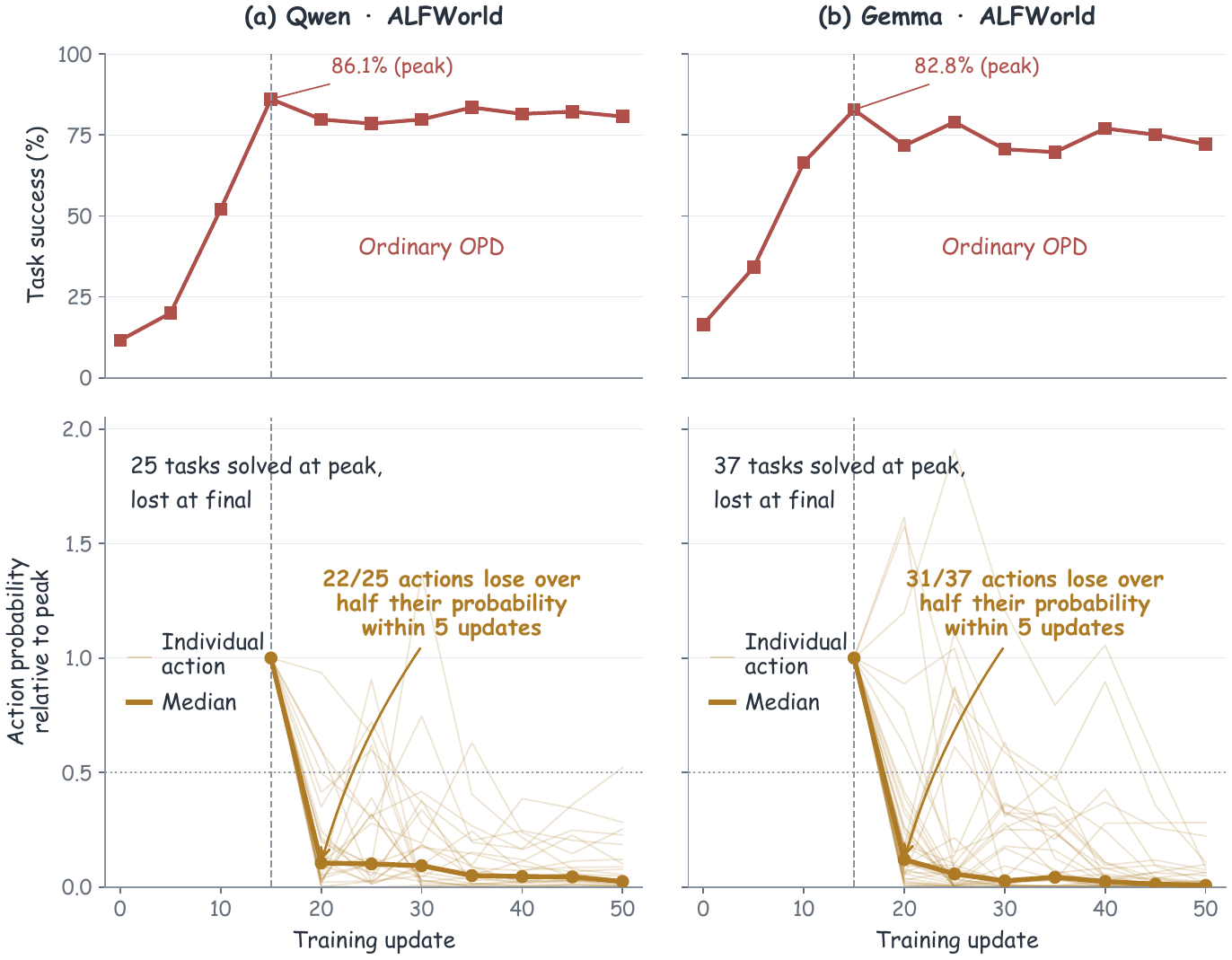}
\caption{\textbf{Action erosion accompanies post-peak regression.}
Top: ordinary OPD's success on Qwen/Gemma ALFWorld.
Bottom: As task success declines, the majority of tracked actions lose over half their probability within five updates of the peak. Action halving occurs at or before the corresponding task's first failure in 96\%/92\% of lost tasks, linking decision-level erosion to task-level regression.}
\label{fig:erosion_dynamics}
    \vspace{-10pt}
\end{figure}

\paragraph{Decision-level erosion translates into task-level regression.}
Does this erosion explain why tasks are lost? For the 25 Qwen tasks
solved at the peak but lost by the end, we track the action on each
successful peak trajectory whose probability falls most. We find that these actions
are almost always teacher-disfavored (92\%) and erode quickly: 22 of 25
halve within five updates, at or before the checkpoint where the task is
first lost in 96\% of tasks (76\% at exactly that checkpoint;
\autoref{fig:erosion_dynamics}). Gemma shows the same pattern across 37
lost tasks. Tasks are thus lost when \textbf{continued imitation erodes
a decisive choice the student had gotten right}.

In tasks that remain solved, erosion is milder (median 7-fold versus
41-fold) and rarely decisive, since an alternative route usually survives.
Repair also continues after the peak (Qwen gains 85.7 task-equivalents)
but is outpaced by erosion (100.7 lost), thus producing the regression.

\subsection{A Stylized Model of Repair and Erosion}
\label{sec:overshoot_theory}

We formalize how simultaneous repair and erosion yield an above-teacher
intermediate policy even as the imitation loss decreases. Consider an
episode requiring two correct binary decisions, with student success
probabilities $p_i=\sigma(z_i)$, teacher probabilities $q_i$, and episode
success $J=p_1p_2$ versus $J_T=q_1q_2$. Under forward-KL gradient flow,
\begin{equation}
 \dot p_i=\lambda_i p_i(1-p_i)(q_i-p_i),\qquad \lambda_i>0,
 \label{eq:flow}
\end{equation}
where $\lambda_i$ is each decision's effective update rate. Let the student
start with one strength, $p_1(0)>q_1$, and one weakness, $p_2(0)<q_2$,
with $J(0)<J_T$.

\begin{proposition}[Transient superiority]
\label{prop:overshoot}
If $\lambda_2q_2(1-q_2)>\lambda_1q_1(1-q_1)$, the student exceeds the
teacher at a finite time, reaches an above-teacher maximum, and then
decreases toward $J_T$, while the total forward-KL loss strictly decreases.
\end{proposition}

When the weak decision is repaired faster than the strength erodes, the
student briefly holds both; continued imitation then removes the strength.
With $q_i=0.83$ (teacher success 68.9\%, as for Gemma on ALFWorld), $p(0)=(0.99,0.10)$ and $\lambda=(0.025,1)$, success rises from 9.9\% to 82.0\% before settling at the teacher's level.
Here $\lambda$ is chosen for illustration rather than fitted. We defer proofs to \autoref{app:proofs}.

\subsection{Teacher Reliability Predicts Where Gains Are Fragile}
\label{sec:fragility}

To locate post-peak losses, we group tasks by how often
the teacher solves them. Losses concentrate where the teacher is
unreliable: on tasks it never solves, success falls from the peak to the
final checkpoint by 27.4 points for Qwen and 22.1 for
Gemma, while tasks it always solves stay near their
peak (\autoref{fig:teacher_cohorts});
this holds in all five ordinary-OPD runs. Task families also peak at
different times: for Gemma, on ALFWorld, ``Look'' peaks at update
15 and falls from 74.2\% to 28.0\%, ``Clean'' peaks at update 25 and falls to
50.0\%, while ``Cool'' and ``Heat'' peak at update 30 and hold. Thus no single stopping time, and no family-level rule,
retains every gain, motivating decision-level protection.

\section{Success-Preserving On-Policy Distillation (SPOD)}
\label{sec:method}

The preceding analysis motivates protecting successful
student choices from conflicting supervision while retaining
useful teacher guidance. We therefore propose SPOD, a lightweight intervention which
keeps ordinary OPD's student rollouts and teacher scoring
and modifies only the loss at each sampled decision.

\subsection{The SPOD Objective}
\label{sec:spod_objective}

\paragraph{Identifying choices to protect.}
Let $R(\xi)\in\{0,1\}$ indicate whether a student trajectory $\xi$
succeeds, and for an action $a=(a_1,\ldots,a_n)$ at history $s$ define
the sequence log-probabilities
\begin{equation}
\begin{aligned}
 \ell_p &= \sum_{j=1}^{n}\log p_\theta(a_j\mid s,a_{<j}),\\
 \ell_q &= \sum_{j=1}^{n}\log q(a_j\mid s,a_{<j}).
\end{aligned}
\label{eq:seq}
\end{equation}
SPOD protects an action when its trajectory succeeds and the teacher
assigns it lower probability than the student:
\begin{equation}
 g(s,a,\xi)=\mathbf{1}\{R(\xi)=1\ \land\ \ell_q<\ell_p\}.
 \label{eq:gate}
\end{equation}
Success marks behavior worth keeping and relative teacher support marks
conflict, so the gate neither protects disagreements in failed episodes
nor withholds guidance from entire successful trajectories. It is
computed before each update and held fixed during differentiation.

\paragraph{Masking conflicting supervision and countering erosion.}
For gated actions, SPOD removes the distillation term and applies a soft
probability floor,
\begin{equation}
 \ell_{\mathrm{floor}}(s,a)=\tfrac{1}{n}\bigl[\log\tau_0-\ell_p\bigr]_+,
 \label{eq:floor}
\end{equation}
giving the objective
\begin{equation}
 \mathcal L_{\mathrm{SPOD}}=\mathbb E_{\xi,s,a}\!
 \left[(1-g)\,\ell_{\mathrm{KD}}+\lambda\, g\,\ell_{\mathrm{floor}}\right].
 \label{eq:spod}
\end{equation}
The floor raises a protected action only while its probability is below
$\tau_0$ and is inactive above it. We set $\tau_0=0.5$, the smallest value
at which the protected action is guaranteed to remain the student's most
likely action, and $\lambda=1$, unchanged across all settings. Masking
removes the teacher's pull at the decision itself; the floor counters
declines caused by updates elsewhere. All ungated decisions receive
ordinary forward-KL supervision.

\subsection{Masking Makes the Floor Effective}
\label{sec:mask_floor}

A natural alternative is to add the floor to ordinary OPD without
masking. However, this would pit the teacher's
pull against the floor's push, settling the protected action below the
floor. SPOD avoids this by removing the teacher term on protected actions,
leaving the floor unopposed. A
one-decision model makes this precise. Let action 1 succeed and action 0
fail, with student and teacher probabilities $p=\sigma(z)$ and $q$ for
action 1; actions are sampled from the student, and the sampling
distribution and gate are held fixed, as in SPOD.

\begin{proposition}[Masking enables preservation]
\label{prop:floor}
Let $\lambda=1$, $0<q<\tau_0^2$ and $q<p<\tau_0$. Keeping the
distillation term on gated actions and adding the floor gives
$\mathbb{E}[\nabla_z\ell_{\mathrm{add}}]=p^2-q$, with a stable equilibrium
at $p=\sqrt q<\tau_0$. SPOD instead gives
$\mathbb{E}[\nabla_z\ell_{\mathrm{SPOD}}]=-q(1-p)<0$ below the floor and
$(1-p)(p-q)>0$ just above it, so $\tau_0$ is an attracting boundary.
\end{proposition}

For example, with $q=0.16$ and $\tau_0=0.5$, the floor alone settles at a
compromise with the teacher at $p=0.4$, while masking lets the protected
action reach the floor. This matches the ablation: the floor alone or
masking alone underperforms their combination
(\autoref{sec:ablation}). Proofs and an analysis of shared-parameter
effects are in \autoref{app:proofs}.

\paragraph{Implementation and cost.}
SPOD adds negligible cost: it reuses OPD's student and teacher scoring passes, adding only sequence log-probability sums and a scalar gate. Action probabilities include the end-of-action token and are computed before temperature or action constraints. Losses are averaged over episodes, states, then tokens, so long actions do not dominate. Beyond ordinary OPD, SPOD needs only binary episode success: no stored policy, per-family anchor, extra rollout, or teacher judgment are required.

\begin{table*}[!t]
\centering
\small
\setlength{\tabcolsep}{4.1pt}
\begin{tabular}{l ccccccc c c c}
\toprule
& \multicolumn{7}{c}{\textbf{ALFWorld}}
& \textbf{WebShop} & \textbf{ScienceWorld} & \textbf{Overall} \\
\cmidrule(lr){2-8}
\textbf{Method} & \textbf{Pick} & \textbf{Look}
& \textbf{Clean} & \textbf{Heat} & \textbf{Cool}
& \textbf{Pick2} & \textbf{Avg}
& \textbf{Succ.} & \textbf{Succ.} & \textbf{Avg} \\
\midrule
\addlinespace[3pt]
\multicolumn{11}{l}{\textbf{Qwen3.5-2B $\leftarrow$ Qwen3.5-27B}} \\
\addlinespace[3pt]
Teacher 0-Shot                  & 98.3 & 80.6 & 74.7 & 53.0 & 72.5 & 95.1 & 80.0 & 34.1 & 42.4 & 52.2 \\
Student 0-Shot                  & 18.6 & 55.9 & 0.0  & 0.0  & 0.0  & 11.4 & 11.6 & 7.9  & 6.1  & 8.5  \\
\addlinespace[3pt]
Ordinary OPD                    & 99.4 & 77.4 & 71.3 & 67.5 & 69.6 & 94.3 & 80.7 & 38.0 & 41.3 & 53.3 \\
\quad best ckpt. (peak) $^\ast$ & 97.2 & 90.3 & 86.8 & 73.5 & 79.0 & 86.2 & \underline{86.1} & \underline{44.0} & \underline{44.4} & \underline{58.2} \\
\quad stop at eval dip$^\ast$   & 97.2 & 90.3 & 86.8 & 73.5 & 79.0 & 86.2 & \underline{86.1} & \underline{44.0} & \underline{44.4} & \underline{58.2} \\
\quad stop at train dip         & 100.0 & 83.9 & 77.6 & 47.0 & 70.3 & 92.7 & 79.8 & \underline{44.0} & 40.6 & 54.8 \\
\quad retire at teacher level   & 100.0 & 83.9 & 77.6 & 47.0 & 70.3 & 92.7 & 79.8 & 38.4 & 41.3 & 53.2 \\
\addlinespace[3pt]
TCOD                            & 100.0 & 87.1 & 66.0 & 65.8 & 57.9 & 90.2 & 77.9 & 40.1 & 37.8 & 51.9 \\
Guided-OPD                      & 98.8 & 83.8 & 71.2 & 66.6 & 63.7 & 87.8 & 79.2 & 37.2 & 41.3 & 52.6 \\
SAGE-OPD                        & 98.8 & 87.1 & 71.3 & 66.6 & 70.3 & 86.2 & 80.4 & 39.8 & 40.4 & 53.5 \\
\textbf{SPOD (ours)}            & 99.4 & 86.0 & 89.7 & 77.8 & 90.6 & 97.6
  & \textbf{91.0}\gain{10.3} & \textbf{50.3}\gain{12.3} & \textbf{55.9}\gain{14.6} & \textbf{65.7}\gain{12.4} \\
\midrule
\addlinespace[3pt]
\multicolumn{11}{l}{\textbf{Gemma-4-E4B $\leftarrow$ Gemma-4-31B}} \\
\addlinespace[3pt]
Teacher 0-Shot                  & 100.0 & 46.2 & 35.6 & 69.2 & 80.4 & 74.8 & 68.9 & 38.0 & \underline{61.1} & 56.0 \\
Student 0-Shot                  & 45.2  & 20.4 & 5.7  & 2.6  & 6.5  & 12.2 & 16.5 & 27.3 & 22.6 & 22.1 \\
\addlinespace[3pt]
Ordinary OPD                    & 92.7  & 28.0 & 50.0 & 84.6 & 83.3 & 93.5 & 73.7 & 40.5 & 53.9 & 56.0 \\
\quad best ckpt. (peak) $^\ast$ & 97.7  & 62.4 & 74.1 & 84.6 & 76.8 & 94.3 & \underline{82.8} & \underline{44.3} & 53.9 & \underline{60.3} \\
\quad stop at eval dip$^\ast$   & 97.7  & 62.4 & 74.1 & 84.6 & 76.8 & 94.3 & \underline{82.8} & \underline{44.3} & 48.9 & 58.7 \\
\quad stop at train dip         & 97.7  & 62.4 & 74.1 & 84.6 & 76.8 & 94.3 & \underline{82.8} & \underline{44.3} & 48.3 & 58.5 \\
\quad retire at teacher level   & 95.5  & 74.2 & 66.1 & 82.1 & 63.8 & 86.2 & 78.2 & \underline{44.3} & 48.3 & 56.9 \\
\addlinespace[3pt]
TCOD                            & 97.7  & 32.2 & 37.3 & 70.9 & 73.9 & 74.8 & 66.3 & 41.4 & 48.7 & 52.1 \\
Guided-OPD                      & 100.0 & 21.5 & 83.9 & 87.1 & 77.5 & 47.1 & 74.2 & 36.0 & 54.8 & 55.0 \\
SAGE-OPD                        & 97.7 & 37.6 & 74.1 & 72.6 & 79.0 & 73.2 & 75.5 & 40.5 & 52.2 & 56.1 \\
\textbf{SPOD (ours)}            & 100.0 & 57.0 & 92.5 & 87.2 & 86.2 & 96.7
  & \textbf{88.9}\gain{15.2} & \textbf{48.1}\gain{7.6} & \textbf{61.7}\gain{7.8} & \textbf{66.2}\gain{10.2} \\
\bottomrule
\end{tabular}
\caption{Final success rates (\%). ALFWorld is reported per task family;
WebShop and ScienceWorld overall; Overall averages the three benchmarks.
Subscripts indicate SPOD's gain over ordinary OPD. $^\ast$Uses evaluation
results to choose the checkpoint (oracle). SPOD outperforms all compared methods on every benchmark.}
\label{tab:main}
\end{table*}

\section{Experiments}
\label{sec:experiments}
\subsection{Setup}
\paragraph{Models and benchmarks.}
We evaluate two model families, Qwen3.5-2B distilled from Qwen3.5-27B and
Gemma-4-E4B distilled from Gemma-4-31B \citep{qwen35,gemma4}, on three
multi-turn agentic benchmarks: ALFWorld \citep{shridhar2021alfworld}, WebShop \citep{yao2022webshop}, and ScienceWorld \citep{wang2022scienceworld} (we follow common practice
and train on a subset of task types \citep{xi2025agentgymrl,yang2026actmore} chosen by a zero-shot evaluation before \emph{any} training, requiring the teacher to outperform a non-zero student to ensure \emph{distillation is meaningful}). 

\begin{table}[H]
\centering
\vspace{-5pt}
\scriptsize
\setlength{\tabcolsep}{4pt}
\begin{tabular}{llrrr}
\toprule
Benchmark & Domain & Task families & Train & Test \\
\midrule
ALFWorld     & Household tasks     & 6       & 3,553 & 274 \\
WebShop      & Online shopping     & --      & 6,410 & 500 \\
ScienceWorld & Science experiments & 6/30 & 2,880 & 180 \\
\bottomrule
\end{tabular}
\caption{Three benchmarks we adopt and evaluate.}
\label{tab:benchmarks}
\vspace{-10pt}
\end{table}
\paragraph{Comparison methods.}
We compare against three recent agentic OPD methods:
TCOD varies the rollout horizon \citep{wang2026tcod},
Guided-OPD introduces teacher interventions \citep{li2026guidedopd},
and SAGE-OPD selects turns for supervision \citep{zhou2026sageopd}.
Our comparisons focus on preserving gains within
teacher-supervised OPD, with success-conditioned self-imitation
and extensive component ablations testing alternative uses of the same
outcome information.

\paragraph{Controls and ablations.}
We additionally report (1) the student before OPD, (2) ordinary OPD's best
checkpoint, (3) stopping at the first evaluation dip, (4) stopping at the
first training-success dip, and (5) retiring at teacher level. On Qwen
ALFWorld, we also compare (6) a lower learning rate, (7) weight averaging,
(8) reverse KL, (9) teacher-confidence gating, and (10) self-imitation, and
ablate SPOD into (11) masking only, (12) floor only, and (13) episode-level
filtering.

\paragraph{Training and evaluation.}
We follow the standard OPD training protocol with a cosine learning-rate
schedule. Teachers are fixed off-the-shelf models. For ordinary OPD and variants, we use the peak learning rate that works best
in each setting from \{$1\times10^{-5}$, $1\times10^{-6}$, $2\times10^{-6}$\}, further lowered when training diverged. For SPOD (Ours), we use a fixed learning rate of $10^{-5}$, $\tau_0=0.5$, and $\lambda=1$
in all settings. Additional training and evaluation details, including random
seeds, are provided in \autoref{app:protocol}.

\begin{table}[t]
\vspace{-10pt}
\centering\small
\setlength{\tabcolsep}{5pt}
\begin{tabular}{lc}
\toprule
Method & Final (\%) \\
\midrule
Ordinary OPD                        & 80.7 \\
\quad best ckpt.$^\ast$             & 86.1 \\
\midrule
Lower learning rate ($\times0.3$)   & 82.8 \\
Weight averaging (EMA)              & 80.9 \\
Reverse KL                          & 82.5 \\
Reverse KL, lower learning rate     & 77.0 \\
Teacher-confidence gating           & 74.1 \\
Retire at teacher level             & 79.8 \\
OPD + NTP on successes         & 84.4 \\
\midrule
SPOD                                & \textbf{91.0} \\
\bottomrule
\end{tabular}
\caption{\textbf{Alternatives to SPOD.} Final success on Qwen ALFWorld. $^\ast$Selected on evaluation results (oracle).}
\label{tab:alternatives}
\vspace{-10pt}
\end{table}

\subsection{Main Results}
\label{sec:results}

\paragraph{SPOD improves final performance in every setting.}
As shown in \autoref{tab:main}, SPOD achieves the best final success rate
in every setting. Over ordinary OPD, it gains 10.3, 12.3, and 14.6 points
for Qwen and 15.2, 7.6, and 7.8 points for Gemma on ALFWorld, WebShop, and
ScienceWorld, raising the overall average by 12.4 and 10.2 points. SPOD
also surpasses the teacher everywhere, by up to 20.0 points, with a single
untuned configuration.

\paragraph{SPOD surpasses even the best ordinary-OPD checkpoint.}
No ordinary-OPD checkpoint, even one selected on evaluation results,
matches SPOD's final model (e.g., 86.1\% vs.\ 91.0\% for Qwen and 82.8\%
vs.\ 88.9\% for Gemma on ALFWorld). Practical stopping rules recover at
most the peak and often stop early. SPOD thus not only retains the
overshoot but extends it.

\paragraph{Gains concentrate where ordinary OPD regresses.}
The largest gains occur in families that ordinary OPD loses after its
peak, typically where the teacher is weak: Gemma Clean rises from 50.0\%
to 92.5\% and Look from 28.0\% to 57.0\%; Qwen Cool rises from 69.6\% to
90.6\% and Clean from 71.3\% to 89.7\%.

\paragraph{Guidance-based baselines still regress heavily.}
TCOD, Guided-OPD, and SAGE-OPD change when or where the teacher
supervises, but not what it pulls the student away from. They therefore
overshoot and regress as well: Guided-OPD falls from 67.2\% to 54.8\% with Gemma and TCOD from 45.6\% to 37.8\% with Qwen on ScienceWorld.

\paragraph{SPOD can complement other OPD methods.}

\begin{wraptable}{r}{0.55\columnwidth}
\vspace{-10pt}
\centering
\footnotesize
\setlength{\tabcolsep}{3pt}
\begin{tabular}{@{}lr@{}}
\toprule
Method & Final (\%) \\
\midrule
Guided-OPD        & 54.8 \\
SPOD              & 61.7 \\
Guided-OPD + SPOD & \textbf{66.5} \\
\bottomrule
\end{tabular}
\caption{Complementary effect of SPOD.}
\label{tab:combo}
\vspace{-10pt}
\end{wraptable}

Adding SPOD to Guided-OPD on Gemma ScienceWorld raises
final success to 66.5\%, improving on Guided-OPD alone
by 11.7 points and SPOD alone by 4.8 points
(\autoref{tab:combo}), suggesting that SPOD
could serve as a \emph{drop-in addition} to existing agentic OPD methods rather than an alternative to them.

\begin{figure}[!t]
    \centering
    \includegraphics[width=1\linewidth]{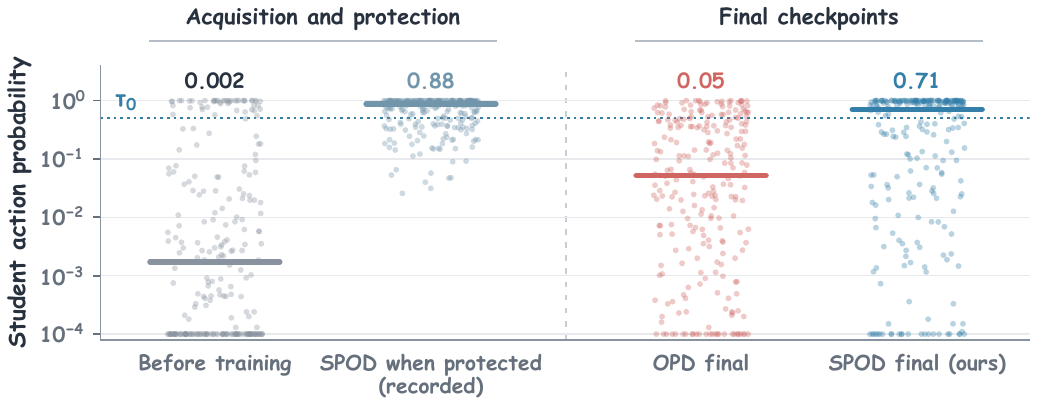}
\caption{\textbf{SPOD keeps successful choices the teacher disfavors.} Each dot is one such action (Gemma, ALFWorld); bars mark medians. Left: the student learns these actions during training. Right: at the final checkpoints, ordinary OPD suppresses them, while SPOD keeps most above $\tau_0$ (dotted line).}
\label{fig:protected}
\vspace{-10pt}
\end{figure}

\begin{figure}[!t]
    \centering
    \includegraphics[width=0.95\linewidth]{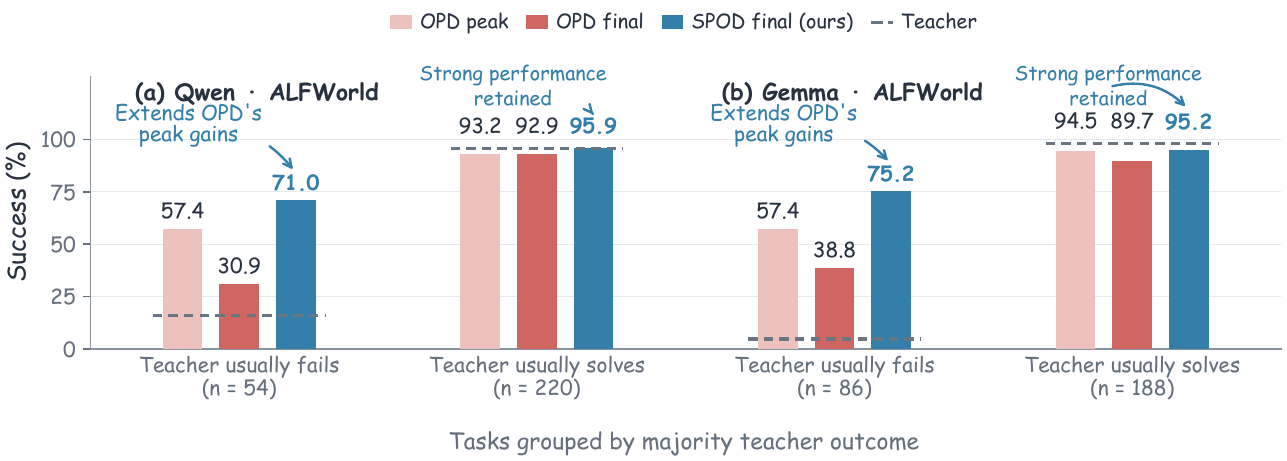}
\caption{\textbf{SPOD preserves and extends the overshoot.} On tasks the teacher usually fails (left bars), ordinary OPD drops from its peak to its final checkpoint, while SPOD ends above that peak. }
\label{fig:cohorts}
    \vspace{-10pt}
\end{figure}

\subsection{SPOD Preserves the Overshoot}
\label{sec:preserve}

\paragraph{Decision level.}
\autoref{fig:protected} tracks the actions SPOD protects: successful choices the teacher disfavors, which erode under ordinary OPD (\autoref{sec:erosion}). The student acquires them during training, with median probability rising from 0.002 before training to 0.88 when protected. At the final checkpoints, ordinary OPD assigns these same actions a median of 0.05, whereas SPOD keeps 0.71, above $\tau_0$, the empirical counterpart of \autoref{fig:overshoot_window}.

\paragraph{Task level.}
Preserving these choices preserves the tasks they solve. On ALFWorld tasks the teacher usually fails, where ordinary OPD's losses concentrate (\autoref{sec:fragility}), ordinary OPD falls from 57.4\% at its peak to 30.9\% (Qwen) and 38.8\% (Gemma), whereas SPOD ends at 71.0\% and 75.2\%, above ordinary OPD's peak (\autoref{fig:cohorts}). On tasks the teacher usually solves, SPOD matches or exceeds ordinary OPD (95.9\% vs.\ 92.9\%; 95.2\% vs.\ 89.7\%), so protection does not stop the student from learning elsewhere.

\begin{table}[!t]
\centering
\scriptsize
\setlength{\tabcolsep}{3pt}
\renewcommand{\arraystretch}{1.08}
\resizebox{\columnwidth}{!}{%
\begin{tabular}{@{}lccrr@{}}
\toprule
Variant & Mask & Floor & Qwen & Gemma \\
\midrule
Ordinary OPD       & --      & --           & 80.7 & 73.7 \\
Floor only         & --      & $\checkmark$ & 84.5 & 72.6 \\
Episode-level mask & episode & --           & 87.5 & 86.3 \\
Step-level mask    & step    & --           & 87.8 & 87.0 \\
SPOD (ours)        & step    & $\checkmark$ & \textbf{91.0} & \textbf{88.9} \\
\bottomrule
\end{tabular}%
}
\caption{\textbf{Masking and the floor work best together.}
Final ALFWorld success (\%).
Episode-level masking removes teacher supervision from every
action in successful episodes; step-level masking removes it
only where the teacher assigns the action lower probability
than the student. The floor applies to these same
teacher-disfavored successful actions.}
\label{tab:ablation}
\vspace{-10pt}
\end{table}

\subsection{Ablation Study and Alternatives}
\label{sec:ablation}

\paragraph{Masking enables effective preservation.}
Table~\ref{tab:ablation} shows that removing conflicting
supervision provides most of SPOD's improvement: masking alone
gains 7.1 points on Qwen and 13.3 on Gemma. Step-level masking
slightly outperforms episode-level masking while retaining
teacher guidance on nonconflicting decisions. The floor alone
provides inconsistent gains, yet adds a further 3.2 and 1.9
points when paired with masking. This matches
\autoref{sec:mask_floor}: with teacher supervision retained,
the floor competes with a conflicting target; masking removes
that opposition, allowing the floor to counter subsequent erosion.
\paragraph{Simpler alternatives do not retain the gains.}
Lower learning rates, weight averaging, reverse KL, and
teacher-based gating or stopping all underperform SPOD
(\autoref{tab:alternatives}). Self-imitation adds a next-token
loss on successful student trajectories while retaining teacher
supervision, reaching 84.4\%, but still considerably below ordinary OPD's peak
(86.1\%) and SPOD (91.0\%). Combined with the masking ablation,
this also shows that simply reinforcing successful behavior is less
effective than removing conflicting supervision and preserving
those choices. SPOD also outperforms tuned GRPO trained on the same
success signal by 5.5 points on average across the three
Qwen benchmarks (\autoref{tab:grpo} \& \autoref{app:grpo}).

\section{Conclusion}
\label{sec:conclusion}
We characterize \emph{distillation overshoot} in multi-turn
agentic OPD: smaller, initially weaker students can substantially
surpass their larger teachers mid-training before regressing.
The teacher initially repairs weak decisions faster than it
erodes successful student choices; regression follows as erosion
takes over. Motivated by this diagnosis, SPOD withholds conflicting
teacher supervision and adds a soft probability floor on successful
actions the teacher disfavors. Across six model--benchmark
settings, SPOD outperforms the teacher,
ordinary OPD, and its best intermediate checkpoints, preserving
and extending the overshoot gains. Results with Guided-OPD
further demonstrate SPOD's potential to complement existing
OPD methods.

\clearpage

\section*{Limitations}
\label{sec:limitations}

\paragraph{Statistical support.}
SPOD, ordinary OPD, alternatives, ablations, and sensitivity
experiments use three training seeds; TCOD, Guided-OPD, and
SAGE-OPD use one each. Every checkpoint is evaluated with
three evaluation seeds. Additional training seeds would
strengthen comparisons with the single-seed baselines.

\paragraph{Experimental scope.}
We evaluate two teacher--student pairs and three text-based
agentic benchmarks. Compute constraints limit our study to
smaller students; larger-scale distillation remains a natural
extension. Within each pair, student and teacher share a
vocabulary, enabling full-vocabulary distillation. Extending
SPOD across tokenizers and to real-world environments would
broaden its applicability.

\paragraph{Replication and coverage.}
SPOD, ordinary OPD, alternatives, ablations, and sensitivity
experiments use three training seeds, while TCOD, Guided-OPD,
and SAGE-OPD use one each. All checkpoints are evaluated
with three evaluation seeds. Additional training seeds would
strengthen comparisons with these methods. Hyperparameter
sensitivity and the alternatives comparison focus on Qwen
ALFWorld, while composition with Guided-OPD is evaluated on
Gemma ScienceWorld. The latter demonstrates a promising
combination; broader evaluation would establish how widely
this complementarity extends.

\paragraph{Mechanism analysis.}
Our detailed decision-level analysis focuses on ALFWorld.
The action-tracing evidence is retrospective, while the
two-decision model isolates repair and erosion using
independent decisions and illustrative update rates.
Extending these probes to other environments and testing
interventions on individual decisions would further
characterize the mechanism.

\paragraph{Outcome supervision.}
SPOD uses binary episode success to identify behavior worth
preserving, without requiring action-level annotations.
This signal does not distinguish each action's contribution
to success. The current gate assumes reliable episode
outcomes; incorporating noisy or graded feedback could
extend preservation to tasks with less structured evaluation.
\bibliography{custom}
\clearpage

\appendix
\section{Ethics Statement}
\label{app:ethics}
Our experiments use existing models and benchmark environments,
with no human participants or collection of personal data.
We identify no ethical concerns specific to this study.

\section{Reproducibility Statement}
\label{app:reproducibility}
We will publicly release all training and evaluation code,
experiment configurations, and trained student checkpoints
to support reproduction of our results and further research
on agentic distillation.

\section{AI Use Statement}
We used AI tools to assist with code development, figure generation,
and manuscript polishing. The authors reviewed and verified the
resulting code, figures, and text and take full responsibility for
the paper's content.

\section{Proofs and Additional Theory}
\label{app:proofs}
\subsection{Proof of Proposition~\ref{prop:overshoot}}
Write $p_i=\sigma(z_i)$, let $a=p_1(0)$ and $b=p_2(0)$,
and define
$L_i=D_{\mathrm{KL}}(\operatorname{Ber}(q_i)
\Vert\operatorname{Ber}(p_i))$. Direct differentiation gives $\partial L_i/\partial z_i=p_i-q_i$. Consequently, under Eq.~\eqref{eq:flow},
\begin{equation}
 \frac{d}{dt}\sum_iL_i=-\sum_i\lambda_i(p_i-q_i)^2<0
 \label{eq:loss_dec}
\end{equation}
at every finite time for the specified nonstationary initialization. Each probability approaches its teacher target monotonically without crossing it. Put $u(t)=p_1(t)-q_1>0$ and $v(t)=q_2-p_2(t)>0$. Both tend to zero, and their local dynamics give
\begin{align}
 u(t)&=C_1e^{-\kappa_1t}(1+o(1)),\\
 v(t)&=C_2e^{-\kappa_2t}(1+o(1)),
\end{align}
where $C_i>0$ and $\kappa_i=\lambda_iq_i(1-q_i)$. To obtain these asymptotics directly, divide each scalar equation by its distance from equilibrium. The resulting logarithmic derivative converges to $-\kappa_i$; its remaining term is integrable because the distance already decays exponentially in a neighborhood of equilibrium. Thus the rescaled distances converge to positive constants.

The performance difference is
\begin{equation}
 J(t)-J_T=q_2u(t)-q_1v(t)-u(t)v(t).
\end{equation}
If $\kappa_2>\kappa_1$, the positive first term dominates for sufficiently large $t$. Hence $J(t)>J_T$ eventually. Substitution of the scalar dynamics also gives
\begin{equation}
 \dot J(t)=-\kappa_1q_2C_1e^{-\kappa_1t}(1+o(1))<0
\end{equation}
for all sufficiently large times. Because $J(0)<J_T$ and $J(t)\to J_T$, continuity ensures a finite crossing and a finite above-teacher maximum. This proves the proposition. The statement guarantees eventual regression; it does not require every point before the maximum to be increasing. If an initially increasing trajectory is desired, the additional condition is
\begin{equation}
 \lambda_2(1-b)(q_2-b)>\lambda_1(1-a)(a-q_1),
\end{equation}
which follows by evaluating $\dot J(0)/J(0)$.

\begin{figure}
    \centering
    \includegraphics[width=1\linewidth]{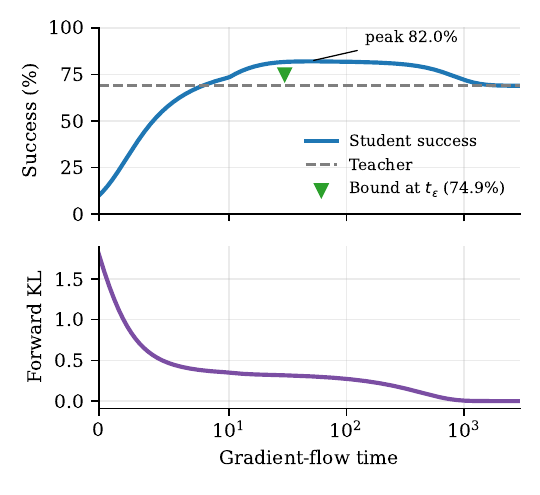}
\caption{\textbf{Constructive overshoot in the two-decision
model} (Proposition~3.1). With the parameters of Section~3.4,
success rises from 9.9\% to a peak of 82.0\%, above the
teacher's 68.9\%, then converges to the teacher's level
while forward-KL loss decreases monotonically. The triangle
marks the analytic lower bound on success at $t_\epsilon$
(74.9\%). Time denotes gradient-flow time, not training updates.}
    \label{fig:theory}
\end{figure}

\paragraph{An explicit finite-time lower bound.}
For $0<\epsilon<q_2-b$, define
\begin{align}
 m_2&=\min\{b(1-b),q_2(1-q_2)\},\\
 t_\epsilon&=\frac{\log((q_2-b)/\epsilon)}{\lambda_2m_2}.
\end{align}
Since $p_2$ increases on $[b,q_2]$, its deficit obeys $\dot v\leq-\lambda_2m_2v$, so $p_2(t_\epsilon)\geq q_2-\epsilon$. Meanwhile
\begin{equation}
 \dot p_1\geq-\lambda_1(a-q_1)/4,
\end{equation}
using $p_1(1-p_1)\leq1/4$ and $p_1-q_1\leq a-q_1$. Whenever the first factor is nonnegative,
\begin{equation}
 J(t_\epsilon)\geq
 \left[a-\frac{\lambda_1(a-q_1)t_\epsilon}{4}\right](q_2-\epsilon).
\end{equation}
With the parameters in \autoref{sec:overshoot_theory} and
$\epsilon=0.05$, we have $m_2=0.09$,
$t_\epsilon\approx29.79$, and a success lower bound of
$0.7490$, approximately 6.0 percentage points above the
teacher. The integrated trajectory in \autoref{fig:theory} peaks at $82.0\%$. Conversely, if both initial student probabilities are at most their teacher targets, the independent dynamics preserve that ordering and cannot overshoot. Local complementarity is essential in this model.

\subsection{Successful Trajectories Enrich Useful Actions}
Let $R$ denote binary eventual success under the rollout
policy $\pi$. Define
$Q^\pi(s,a)=\Pr(R=1\mid s,a)$,
$V^\pi(s)=\Pr(R=1\mid s)$, and
$A^\pi(s,a)=Q^\pi(s,a)-V^\pi(s)$.
For a visited state with $V^\pi(s)>0$, Bayes' rule gives
\begin{align}
 \Pr(a\mid s,R=1)
 &=\frac{\Pr(R=1\mid s,a)\Pr(a\mid s)}{\Pr(R=1\mid s)}\notag\\
 &=\frac{\pi(a\mid s)Q^\pi(s,a)}{V^\pi(s)}.
\end{align}
Subtracting $\pi(a\mid s)$ yields $\pi(a\mid s)A^\pi(s,a)/V^\pi(s)$. Positive-advantage actions are therefore overrepresented among successes. This identity uses the actual rollout policy, including its sampling and action constraints. SPOD's relative-support comparison instead uses raw model sequence probabilities. The roles are different: the outcome selects successful experiences, and the model scores determine where teacher supervision conflicts with student support.

\subsection{Proof of Proposition~\ref{prop:floor}}
For a Bernoulli student, the forward-KL logit derivative is $p-q$. When $p>q$, the gate fires only if action 1 is sampled, an event of probability $p$. Below the floor, the derivative of its negative log probability is $-(1-p)$. Thus the four expected gradients are
\begin{center}
\small
\begin{tabular}{ll}
\toprule
Objective & Expected derivative w.r.t. $z$ \\
\midrule
OPD & $p-q$ \\
Mask only & $(1-p)(p-q)$ \\
Floor only & $p-q-p(1-p)\mathbf{1}_{p<\tau_0}$ \\
SPOD & $(1-p)(p-q)-p(1-p)\mathbf{1}_{p<\tau_0}$ \\
\bottomrule
\end{tabular}
\end{center}
These expectations do not differentiate through the sampling distribution. They average the per-sample gradients produced by the algorithm. For the floor-only objective below $\tau_0$, the expression simplifies to $p^2-q$, negative below $\sqrt q$ and positive above it. Since $q<\tau_0^2$ and $q<\sqrt q$, the equilibrium is inside the active gated region and is stable under gradient descent.

For SPOD below $\tau_0$, the derivative is $-q(1-p)<0$; above $\tau_0$, it is $(1-p)(p-q)>0$. Thus the vector fields point toward the floor from both sides. The hinge is nondifferentiable exactly at the boundary, which is interpreted as an attracting switching threshold; a finite-step optimizer can fluctuate around it. This proves the proposition without claiming an exact stationary probability under arbitrary neural optimization.

\subsection{Local Pressure and Shared-Parameter Coupling}
\label{app:coupling}
For a frozen minibatch with nonnegative weights $w_j$, write $h_i=\nabla_\theta\log p_\theta(a_i\mid s_i)$ and $g_j=\nabla_\theta\ell_j$. A preconditioned step $\Delta\theta=-\eta M\sum_jw_jg_j$, for fixed positive-semidefinite $M$, changes the selected action to first order by
\begin{align}
 \Delta\log p_i={}&-\eta w_i h_i^\top M g_i\notag\\
 &-\eta\sum_{j\ne i}w_jh_i^\top M g_j+O(\eta^2).
 \label{eq:coupled}
\end{align}
The first term is the action's own loss contribution and the second is interference or transfer from other decisions. Masking removes the action's own distillation contribution.
For a gated action below the floor,
$g_i=\lambda\nabla_\theta\ell_{\mathrm{floor},i}
=-\lambda h_i/n_i$, so its own contribution becomes
\begin{equation}
 \frac{\eta w_i\lambda}{n_i}h_i^\top M h_i\geq0.
\end{equation}
Cross-state terms remain. This is the precise local preservation effect; no claim of a universal hard probability guarantee is needed. For an actual optimizer step, including Adam with momentum, the general first-order identity is $\Delta\log p_i=h_i^\top\Delta\theta+O(\|\Delta\theta\|^2)$.

\paragraph{A sequence comparison is not an exact gradient sign.}
Even for a single softmax token under unpreconditioned gradient flow,
\begin{equation}
 \frac{d}{dt}\log p_a=-(p_a-q_a)+\sum_v p_v(p_v-q_v).
\end{equation}
For $p=(0.40,0.59,0.01)$ and $q=(0.39,0.31,0.30)$, the first token has $p_a>q_a$ yet its log probability increases, because the normalization term dominates. Multi-token actions introduce additional coupling. Equation~\eqref{eq:logit} is exact at the logit level; Eq.~\eqref{eq:gate} is a low-cost relative-support selector whose usefulness is tested by the fixed-state probes and ablations.

\begin{figure}[t]
\small
\hrule\vspace{4pt}
\textbf{SPOD training update}\par
\begin{enumerate}[leftmargin=1.5em,nosep]
 \item Roll out the current student on a batch of training tasks; record each episode's binary success.
 \item Sample up to eight decisions per episode and obtain teacher token distributions at those contexts.
 \item Compute $\ell_p$, $\ell_q$, and the detached gate in Eq.~\eqref{eq:gate}.
 \item Use the floor for gated decisions and forward KL for the others; average and update the student.
\end{enumerate}
\vspace{4pt}\hrule
\caption{SPOD uses fresh on-policy data at every update. Evaluation outcomes are never used by the gate.}
\label{fig:algorithm}
\end{figure}

\section{Additional Experiment Details}
\subsection{Additional Implementation Details}
\label{app:protocol}
\paragraph{Task sets and metrics.}
ALFWorld is evaluated on 274 tasks (140 seen, 134 unseen) spanning six
families: Pick (59), Look (31), Clean (58), Heat (39), Cool (46), and
Pick2 (41). The reported average is the task-weighted success rate over
all 274 tasks. WebShop is evaluated on 500 held-out shopping
instructions, and the metric is full task success rather than the
partial reward score. ScienceWorld is evaluated on six task families
(inclined-plane angle determination, friction with named and with
unnamed surfaces, melting-point measurement with known and with unknown
substances, and finding a non-living thing), with 30 held-out instances
each (180 in total).

\paragraph{Training protocol.}
Within our computational budget, we prioritize multi-seed
replication for SPOD, ordinary OPD, and the controlled
alternatives and ablations. Hence, we use
\textbf{three training seeds} for SPOD and ordinary OPD
in \autoref{tab:main}, all alternatives in
\autoref{tab:alternatives}, all ablation variants in
\autoref{tab:ablation}, and all hyperparameter sensitivity experiments in \autoref{tab:sensitivity}, as well as the corresponding figures.
TCOD, Guided-OPD, and SAGE-OPD are each trained with one seed.
For teacher scoring, we format the student's interaction
history using the teacher's native chat template.
Within each model family, the student and teacher share
the same vocabulary and token IDs, enabling full-vocabulary
distillation.

\paragraph{Evaluation protocol.}
We evaluate every checkpoint with three evaluation seeds
and report the mean performance across these evaluations.

\paragraph{Optimization.}
All runs use three warmup updates to the peak learning rate,
followed by cosine decay to one tenth of the peak, with
gradient norm clipped at 1.0. SPOD uses a single fixed peak
learning rate of $10^{-5}$ across all settings, without
learning-rate tuning. Learning-rate selection is applied
only to ordinary OPD and the compared methods: we choose
their best-performing peak learning rate from
$\{10^{-5}, 2\times10^{-6}, 10^{-6}\}$
(\autoref{sec:experiments}). Specifically, ordinary OPD
on Qwen WebShop uses $2\times10^{-6}$, while SAGE-OPD uses
$10^{-6}$, matching its original paper.

\subsection{Sensitivity to $\tau_0$ and $\lambda$}
\label{app:sensitivity}
SPOD has two hyperparameters: the floor level $\tau_0$ and
the floor weight $\lambda$. All main results use fixed
defaults of $\tau_0=0.5$ and $\lambda=1$, \textbf{without
task-specific tuning}. We choose $\tau_0=0.5$ because an
action at or above this probability is at least as likely
as any alternative. Setting $\lambda=1$ gives the floor
the same coefficient as the distillation term it replaces
on protected decisions.

\autoref{tab:sensitivity} varies one hyperparameter at a
time on Qwen ALFWorld, holding everything else fixed.
Every tested setting outperforms ordinary OPD (80.7\%);
all except $\lambda=0.5$ also exceed its best checkpoint
(86.1\%) and the mask-only variant (87.8\%).
The default $\lambda=1$ performs best among the tested
weights, while increasing $\tau_0$ to 0.7 raises final
success to 93.8\%, 2.8 points above the main reported results. Thus, the main results use a shared
default rather than the sweep's best-performing configuration.

\begin{table}[h]
\centering\small
\setlength{\tabcolsep}{7pt}
\begin{tabular}{lccc}
\toprule
& \multicolumn{3}{c}{Final success (\%)} \\
\cmidrule(lr){2-4}
$\tau_0$ ($\lambda=1$)   & 0.3 & \textbf{0.5} & 0.7 \\
                       & 89.4 & \textbf{91.0} & 93.8 \\
\addlinespace[3pt]
$\lambda$ ($\tau_0=0.5$) & 0.5 & \textbf{1} & 2 \\
                       & 84.4 & \textbf{91.0} & 89.4 \\
\bottomrule
\end{tabular}
\caption{\textbf{Sensitivity around fixed defaults.}
Final success on Qwen ALFWorld when varying one
hyperparameter at a time. Bold marks the defaults used
for all main results. Ordinary OPD reaches 80.7\%
at its final checkpoint and 86.1\% at its best checkpoint.}
\label{tab:sensitivity}
\end{table}

\subsection{Definitions of controls, alternatives, and ablations}
\label{app:controls}
All variants use the ordinary-OPD rollout protocol, teacher-scoring
procedure, and evaluation protocol described in
\autoref{sec:analysis} and \autoref{app:protocol}.
Each changes only the component specified below.

\paragraph{Checkpoint selection and stopping (\autoref{tab:main}).}
\emph{Best checkpoint} selects the checkpoint with the highest
evaluation success; \emph{stop at eval dip} selects the checkpoint
immediately before the first decline. Both use evaluation results
and are marked as oracles. \emph{Stop at train dip} stops at the
first decline in training-rollout success between consecutive
five-update windows. \emph{Retire at teacher level} stops once
the student's success over the last five training updates reaches
the teacher's training-rollout success rate, using a teacher-level
criterion inspired by RetireOPD~\citep{yu2026retireopd}.
The latter two rules use only training-time information and report
the checkpoint at the corresponding milestone.

\paragraph{Alternatives (\autoref{tab:alternatives}).}
\emph{Lower learning rate} multiplies the peak and final learning
rates by $0.3$, retaining the schedule shape.
\emph{Weight averaging} evaluates an exponential moving average
of model weights with decay $0.9$ per update.
\emph{Reverse KL} replaces token-level forward KL with
$D_{\mathrm{KL}}(p_\theta\Vert q)$ over the full vocabulary;
its lower-learning-rate variant also scales the rates by $0.3$.
\emph{Teacher-confidence gating} removes distillation on actions
whose mean teacher token entropy exceeds the previous update's
median, selecting supervision by teacher uncertainty.
\emph{OPD + NTP on successes} adds token-averaged negative
log-likelihood of the student's sampled actions from successful
episodes to ordinary OPD. It retains teacher supervision
everywhere and uses the same binary success information as SPOD,
testing whether directly reinforcing successful behavior
suffices to retain the gains.

\paragraph{Ablations (\autoref{tab:ablation}).}
\emph{Floor only} retains forward KL everywhere and adds the
soft floor in \eqref{eq:floor} on actions selected by the
gate in \eqref{eq:gate}, isolating preservation without removing
conflicting supervision.
\emph{Episode-level mask} removes distillation from every action
in a successful episode, with no floor.
\emph{Step-level mask} removes distillation only from successful
actions the teacher disfavors, as selected by the same gate,
with no floor.
SPOD combines this selective mask with the floor.
All floor terms use $\tau_0=0.5$ and $\lambda=1$.
These controls separate the effects of removing teacher
supervision, selecting where to remove it, and countering
subsequent erosion.

\subsection{Distillation overshoot with reverse KL}
\label{app:reverse_overshoot}
To test whether overshoot is specific to forward KL,
we evaluate reverse-KL OPD on Qwen ALFWorld
(\autoref{tab:rkl}). Both tested configurations surpass
the teacher (80.0\%) during training and subsequently
regress. The better-performing configuration peaks at
84.7\% and finishes at 82.5\%, still 8.5 points below
SPOD. These results show that overshoot extends beyond
forward KL and that changing the KL direction alone
does not eliminate post-peak regression.
\begin{table}[h]
\centering\small
\setlength{\tabcolsep}{6pt}
\begin{tabular}{lcc}
\toprule
Objective & Peak (\%) & Final (\%) \\
\midrule
Forward KL (ordinary OPD)        & 86.1 & 80.7 \\
Reverse KL                        & 84.7 & 82.5 \\
Reverse KL, lower learning rate   & 83.5 & 77.0 \\
SPOD (forward KL)                 & 92.0 & \textbf{91.0} \\
\bottomrule
\end{tabular}
\caption{\textbf{Overshoot and regression also occur under reverse KL.}
Peak and final success on Qwen ALFWorld (teacher: 80.0\%).}
\label{tab:rkl}
\end{table}

\begin{table}[h]
\centering
\small
\setlength{\tabcolsep}{5pt}
\begin{tabular}{@{}lrr@{}}
\toprule
Benchmark & GRPO & SPOD (ours) \\
\midrule
ALFWorld     & 85.2 & \textbf{91.0}\;(+5.8) \\
WebShop      & 42.8 & \textbf{50.3}\;(+7.5) \\
ScienceWorld & 52.8 & \textbf{55.9}\;(+3.1) \\
\midrule
Average      & 60.3 & \textbf{65.7}\;(+5.5) \\
\bottomrule
\end{tabular}
\caption{\textbf{SPOD outperforms tuned GRPO.}
Final success (\%) on Qwen3.5-2B, averaged over three
seeds. Parentheses show percentage-point gains over GRPO;
the average gain is computed before rounding.}
\label{tab:grpo}
\end{table}

\subsection{Comparison with outcome-based RL}
\label{app:grpo}
We additionally compare with tuned GRPO on Qwen3.5-2B
(\autoref{tab:grpo}). SPOD achieves higher final success
on all three benchmarks, with an average gain of 5.5 points while using approximately one-quarter of GRPO's training GPU-hours,
extending its advantage beyond distillation baselines.

\end{document}